\documentclass[letterpaper]{article}
\usepackage{amsmath}
\usepackage{graphicx}
\usepackage[margin=2.2cm]{geometry}
\usepackage{url}

\title{PLS in the Mirror of Self-Attention}
\author{
  Jiangsheng (Jason) You\\
  \texttt{jshyou@gmail.com}
}

\begin{document}
\maketitle

\begin{abstract}
\noindent
This note provides an interesting observation on casting partial least square (PLS) as a linearized self-attention so that PLS may be studied within the neural network paradigm. On the other hand, the dimensionality reduction and selection of predictors in PLS may indicate that self-attention includes certain degree of dimensionality normalization toward improved learning.
\end{abstract}

\section{ Introduction on classical PLS}
Let $X$ be an $n\times m$ matrix and $Y$ be an $n\times p$ matrix to represent a pair of observations of independent and dependent variables. In the partial least square (PLS) model, $X$ and $Y$ are called the multivariate predictor and response, respectively, and they can be expressed as
\begin{equation} \label{pls_base}
\begin{aligned}
	X=TP^T+E\\
	Y=UQ^T+F
\end{aligned}
\end{equation}
here $T$ and $U$ are $n\times l$ matrices representing the projections of $X$ and $Y$ by orthogonal $m\times l$ matrix $P$ and $p\times l$ matrix $Q$, respectively, $E$ and $F$ are error terms. The decomposition of $X$ and $Y$ in equation \eqref{pls_base} is made to maximize the cross-covariance between $T$ and $U$ for orthogonal matrices $P$ and $Q$ over all observations of $X$ and $Y$. Let $S=\{(X,Y) \}$ represent all available observations with zero means and $Tr(\cdot)$ denote the trace operator for matrix, then $T$ and $U$ may be solved by finding the solution of the constrained maximization problem
\begin{equation} \label{pls_cov}
		\mathop{\mathrm{\bf argmax}}_{P, Q}\sum_{(X, Y)\in S}Tr[(XP)^TYQ] \textrm{ subject to } P^TP=I \textrm{ and } Q^TQ=I
\end{equation}
Notice that the summation and trace operator used in the definition of \eqref{pls_cov} is to make the maximization general to cover both cases of $n=1$ and $n>1$, this is different from the convention to use $X$ and $Y$ to represent all the observations for $n=1$ in the literature. From the solution of \eqref{pls_cov}, we obtain the score matrices as follows
\begin{equation} \label{pls_score}
T=XP \textrm{ and } U=YQ
\end{equation}
The goal of PLS is to predict the responses from predictors, for this purpose two assumptions are made: 1) $T$ are good predictors of $Y$; and 2) a linear relationship between $T$ and $U$ exists and has the following form
\begin{equation} \label{pls_diag}
	U=TD+H
\end{equation}
where $D$ is a diagonal matrix and $H$ is the error term. Originally, PLS was developed to solve the problem of high collinearity in observed data from multiple industries, see \cite{Herve10, Trygg2002} for a review on this topic and \cite{Herman66, Herman85, Herve03} for early works. Compared with principal component analysis (PCA), PLS finds the components that can provide a good prediction other than the major components. Based on the second assumption, the PLS prediction can be expressed as 
\begin{equation} \label{pls_predict}
	Y=XPDQ^T + R
\end{equation}
Here $R$ is the regression error term. In the implementation of sklearn, $PDQ^T$ and $R$ are named as coefficients and interception of the linear approximation, respectively. There have been many algorithms to solve \eqref{pls_cov} and \eqref{pls_score} but the nonlinear iterative partial least squares (NIPALS) algorithm is the most popular one with a variety of different flavors, see more algorithmic reviews from \cite{Herve10, Trygg2002}.

\section{Architecture of Self-Attention and Transformer}
Attention-based Transformer has quickly become the most effective network architecture for nature language processing (NLP) since its debut in 2017 \cite{Vaswani17}. Here we provide the mathematical details of the matrix operation from the input to its transformed data. Let $W_Q$, $W_K$ and $W_V$ be the weights for query, key and value matrices $Q$, $K$ and $V$, accordingly. Note that $W_Q$, $W_K$ and $W_V$ have the dimension of $m\times l$ as the same dimension of loading matrix $T$, then we have
\begin{equation}
\label{att_qkv}
	Q=XW_Q, \textrm{ } K=XW_K \textrm{ and } V=XW_V
\end{equation}
Notice that three matrices $Q$, $K$ and $V$ are the projections of $X$ to three subspaces possibly in the lower dimension if $l<p$, then the self-attention is defined as
\begin{equation}
\label{att_mapping}
	\mathop{\mathrm{Attention}}(Q, K, V)=\mathop{\mathrm{softmax}}(\frac{QK^T}{\sqrt{l}})V
\end{equation}
Here the softmax operation is applied per-row of the $QK^T$ matrix, which is of size $n \times n$. We rewrite \eqref{att_mapping} as a function of $X$, $W_Q$, $W_K$ and $W_V$ as
\begin{equation}
	\label{att_func}
	\mathop{\mathrm{Attention}}(X;W_Q, W_K, W_V)=\mathop{\mathrm{softmax}}(\frac{XW_QW_K^TX^T}{\sqrt{l}})XW_V
\end{equation}
The encoder of original Transformer architecture actually used a residual transformation followed with a layer normalization step
\begin{equation}
	\label{att_layer_norm}
	X_f=\mathrm{LayerNorm}(\mathrm{Attention}(X;W_Q, W_K, W_V)+X)
\end{equation}

\noindent
Hereafter $X_f$ will be the feature after the encoder steps. Transformer architecture includes two steps, one is encoder and the other is decoder. The output from encoder can be regarded as the feature of $X$ and is plugged in the encoder-decoder block to mix with the attention output from target $Y$. For the language translation, or the above mentioned multivariate prediction, the decoder layer is illustrated in the following picture
\begin{figure}[!h]
	\centering
	\includegraphics[width=0.4\linewidth]{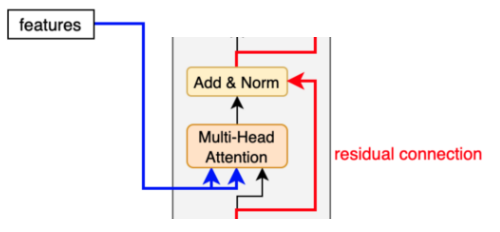}
	\caption{Decoder Block}
	\label{fig:encoder-decoder}
\end{figure}

\noindent
The decoder block is symbolically equivalent to the regression step \eqref{pls_predict} and the corresponding query, key and value matrices are calculated by the following definition
\begin{equation}
	\label{att_reg}
	Q=YW_Q, \textrm{ } K=X_fW_K \textrm{ and } V=X_fW_V
\end{equation}
Here $X_f$ will stand for the features from after encoder of source data and $Y$ represents the features from the attention of target data, then the source-target attention is expressed as
\begin{equation}
	\label{att_func_mix}
	\mathop{\mathrm{Attention}}(X_f, Y;\, W_Q, W_K, W_V)=\mathop{\mathrm{softmax}}(\frac{YW_QW_K^TX^T_f}{\sqrt{l}})X_fW_V
\end{equation}

\noindent
In both encoder and decoder blocks, there is a feed-forward fully-connected layer after the self-attention to further transform the features. The expression in the original paper on Transformer architecture includes two linear transformations as follow
\begin{equation}
	\label{transformer_ffn}
	FFN(X) = \mathrm{max}(0, XW_1+B_1)W_2+B_2
\end{equation}

\section{PLS as a regression over transformed data}
Considering the regression through transformed data through attention and feedforward network in transformer architecture, we discuss a procedure to reformulate PLS to a regression instead of cross covariance minimization. To explicitly express the regression between predictors and responses, we may define the cost function as the squared errors between the predictions and observations of $Y$
\begin{equation}
	\label{pls_mse}
	\mathop{\mathrm{\bf argmin}}_{P, D, Q}\frac{1}{2}\sum_{(X, Y)\in S}|XPD-YQ|^2 \textrm{ subject to } P^TP=I \textrm{ and } Q^TQ=I
\end{equation}
The benefits of \eqref{pls_mse} include both the latent variable extraction and the regression in an explicit way, e.g., $XP=T$ is the projection of predictor $X$ and $TDQ^T$ is the transformation to predict the response $Y$.

To balance the accuracy of latent variable extraction and the regression of responses, we may add a parameterized reconstruction error for $X$ and $Y$ in \eqref{pls_mse} to study the following loss function
\begin{equation}
	\label{pls_loss}
	L(P, Q, D; \alpha, \beta) = \frac{1}{2}[|XPD-YQ|^2 + \alpha |XPP^T-X|^2 + \beta |YQQ^T-Y|^2]
\end{equation}
With the modified cost function \eqref{pls_loss}, we aim to find the solution to a more general minimization problem
\begin{equation}
	\label{pls_model_2}
	\mathop{\mathrm{\bf argmin}}_{P, Q, D}\sum_{(X, Y)\in S}L(P, Q, D; \alpha, \beta) \textrm{ subject to } P^TP=I \textrm{ and } Q^TQ=I
\end{equation}
The restriction on $D$ to be a diagonal matrix may be loosened to be a general matrix with considering certain constraints defined by the problem under consideration. The recent work \cite{You2025} provided a similar schema to reformulate the probabilistic model to a regression. Because of inner product expressions in \eqref{pls_mse} and \eqref{pls_loss}, they can be solved by gradient descent methods.

\vspace{3mm}
\noindent
\textit{Here we use the simple case of one-dimensional target $Y$ to explain the relationship between \eqref{pls_mse} and \eqref{pls_cov}. Mathematically, to minimize $|WX-Y|^2$ is actually equivalent to maximize $(WX)\cdot Y$, indicating that \eqref{pls_mse} may provide more flexibility for PLS to allow non-orthogonal transformations and nonlinear activations.}

\bibliographystyle{plain} 
\bibliography{refs} 

\end{document}